\documentclass[11pt,letterpaper]{mystyle}
\usepackage[comma,authoryear,compress]{natbib}
\hypersetup{
  colorlinks=true,
  citecolor=YaleBlue,
  pdftitle={Omni-Decision: Evidence-Ledger Planning for Omni-Modal Agents},
  pdfauthor={Ming Ma, Yi Zhu, Yiran Zhong, Feida Zhu, Yuhao Wang, Junhan Shi, Lingrui Mei, Tianming Yang, Steven Hoi}
}
\usepackage{eso-pic}
\AddToShipoutPictureFG*{%
  \AtPageUpperLeft{%
    \put(\LenToUnit{1.9cm},\LenToUnit{-1.7cm}){%
      \includegraphics[height=0.9cm]{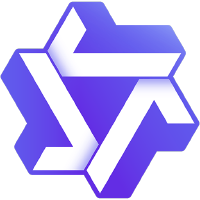}%
    }%
  }%
}

\usepackage[utf8]{inputenc}
\usepackage[T1]{fontenc}
\usepackage{algorithm}
\usepackage{algpseudocode}
\usepackage{hyperref}
\usepackage{url}
\usepackage{booktabs}
\usepackage{multirow}
\usepackage{array}
\usepackage{amsfonts}
\usepackage{amsmath}
\usepackage{amssymb}
\usepackage{nicefrac}
\usepackage{microtype}
\usepackage{xcolor}
\usepackage{colortbl}
\usepackage{graphicx}
\usepackage{fvextra}
\usepackage{enumitem}
\tcbuselibrary{skins,breakable}

\newcommand{\abstractlinkicon}[1]{%
  \makebox[1.15em]{\raisebox{-0.2em}{#1}}\enspace
}

\DefineVerbatimEnvironment{PromptVerbatim}{Verbatim}{
  fontsize=\scriptsize,
  breaklines=true,
  breakanywhere=true,
  breaksymbolleft={},
  breaksymbolright={}
}

\newtcolorbox{prompttemplatebox}[1]{
  enhanced,
  breakable,
  colback=white,
  colframe=black!40,
  colbacktitle=white,
  coltitle=black,
  boxrule=0.5pt,
  arc=2pt,
  left=5pt,
  right=5pt,
  top=5pt,
  bottom=5pt,
  toptitle=2pt,
  bottomtitle=2pt,
  title={\quad \textsc{Prompt Template: #1}},
  fonttitle=\rmfamily\bfseries\small,
  fontupper=\rmfamily\scriptsize
}

\title{Omni-Decision: Evidence-Ledger Planning for Omni-Modal Agents}
\runningtitle{Omni-Decision: Evidence-Ledger Planning for Omni-Modal Agents}
\renewcommand\Affilfont{\centering\normalfont\fontsize{10}{14}\selectfont}
\makeatletter
\renewcommand\AB@affilsepx{\protect\\\protect\Affilfont}
\makeatother
\author[ ]{%
  Ming Ma$^{1,2}$, Yi Zhu$^{3,*}$, Yiran Zhong$^{3,*}$, Feida Zhu$^3$, Yuhao Wang$^4$,\\
  Junhan Shi$^5$, Lingrui Mei$^6$, Tianming Yang$^1$, Steven Hoi$^3$
}
\affil[1]{Institute of Neuroscience, Chinese Academy of Sciences}
\affil[2]{University of Chinese Academy of Sciences}
\affil[3]{Tongyi Lab, Alibaba Group}
\affil[4]{Shanghai Jiao Tong University}
\affil[5]{Tsinghua University}
\affil[6]{Institute of Computing Technology}
\affil[ ]{\texttt{mam2022@ion.ac.cn}, \texttt{zhu.yee@outlook.com}, \texttt{zhongyiran@gmail.com}}
\correspondingauthor{Yi Zhu (\href{mailto:zhu.yee@outlook.com}{zhu.yee@outlook.com}); Yiran Zhong (\href{mailto:zhongyiran@gmail.com}{zhongyiran@gmail.com}). $^{*}$ Corresponding authors.}

\begin{document}
\begin{abstract}
Omni-modal agents must seek evidence across video, audio, web pages, and computation to answer questions. Their main bottleneck is planning: noisy multimodal observations accumulate in conversation history and disrupt later decisions, while multimodal models have limited capacity for multi-step planning. Controlled backend replacements support this diagnosis: replacing the planner causes a much larger performance loss than replacing the perception backend. We present Omni-Decision, an omni-modal agent built on \emph{evidence-ledger planning}: it replaces the growing dialogue history with an explicit evidence ledger that records what evidence is still missing, what has been confirmed, and where records conflict. A critic reads each noisy observation and passes only the usable content to the ledger, discarding the rest, so the planner works from a compact context throughout the task. Each run records the state, action, and verdict at every step, and supervised fine-tuning and decision-level reinforcement learning on these trajectories further improve the planner. Omni-Decision achieves state-of-the-art accuracy of 81.4\% on OmniGAIA at approximately 43\% of Gemini-3.1-Pro's cost per question, and 65.0\% on WorldSense long-video understanding, level with the strongest end-to-end model.

\vspace{1mm}
\noindent\abstractlinkicon{\includegraphics[height=1.1em]{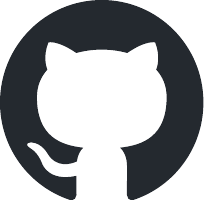}}\textbf{Code:} \url{https://github.com/MBJinX/OmniDecision}\par
\noindent\abstractlinkicon{\includegraphics[height=1.1em]{figures/tongyi-logo.png}}\textbf{Project page:} \url{https://mbjinx.github.io/omnidecision_web/}
\end{abstract}

\maketitle
\vspace{3mm}

\section{Introduction}
\label{sec:introduction}

Omni-modal question answering is moving from closed-form perceptual understanding \citep{wuLongVideoBenchBenchmarkLongcontext2024,fu2024video} toward evidence-seeking tasks \citep{hongWorldSenseEvaluatingRealworld2025,liOmniGAIANativeOmniModal2026}. Given a question grounded in heterogeneous media, a system may need to locate evidence in video, audio, or images, complete missing attributes from the web, and compute derived values before answering \citep{nakanoWebGPTBrowserassistedQuestionanswering2022,renVideoRAGRetrievalAugmentedGeneration2025,yuLongVidSearchAgenticBenchmark2026,liOmniGAIANativeOmniModal2026}. A common approach combines multimodal models with agent frameworks so that evidence can be collected through multi-step tool interaction \citep{yaoReActSynergizingReasoning2023,schickToolformerLanguageModels2023a,taoOmniAgentAudioGuidedActive2025}.

Such systems have two central weaknesses. First, multimodal observations contain substantial noise. A browser returns a full page of mixed text, while video perception produces long descriptions with some irrelevant details. When these observations accumulate in dialogue history, they directly disrupt the multimodal model's planning \citep{lindenbauerComplexityTrap2025,kangACONOptimizingContext2025}: the planner struggles to extract the key information from the noisy context, and it tends to converge prematurely or take unsupported actions.

Second, the capability gap of current multimodal models lies on the planning side. Even when perception provides a clear observation, the model remains limited at multi-step reasoning, requirement tracking, and conflict resolution, while the gap on the perception side is much smaller. Two independent findings support this judgment: coding agents without native audio-video perception can perform comparably to native models on omni-modal benchmarks \citep{chenSandboxedCodingAgents2026}, and the performance difference between agent harnesses for the same model can exceed the difference between models \citep{zhangStopComparingAgents2026}. Our controlled backend replacements turn this judgment into an attribution in the omni-modal domain: replacing the planner causes a much larger performance loss than replacing the perception backend (Section~\ref{sec:experiments-ablation}).

\begin{figure}[t]
  \centering
  \includegraphics[width=\linewidth]{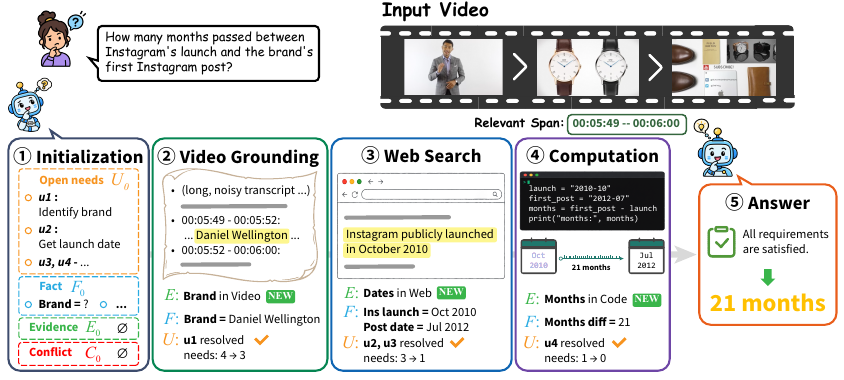}
  \caption{Omni-Decision on an OmniGAIA task. Each tool call is digested into ledger updates that close open evidence needs; once all needs are closed, the system answers against the recorded evidence.}
  \label{fig:omnidecision-main-flow}
\end{figure}

We present Omni-Decision, an omni-modal agent that addresses both weaknesses with \emph{evidence-ledger planning}. Instead of letting observations pile up in dialogue history, each task maintains an evidence ledger of open needs, confirmed evidence, and unresolved conflicts; a critic reads each observation and keeps only the usable content, which a deterministic reducer commits to the ledger (Figure~\ref{fig:omnidecision-main-flow}). Requirement tracking and conflict resolution are thus carried by the ledger, and the planner no longer relies on its own context to remember which evidence is still missing and which records contradict each other. It plans over a compact, verified state, and every step leaves a recorded state, action, and verdict that later post-trains the planner.

Our contributions are:
\begin{itemize}[leftmargin=*,noitemsep,topsep=2pt]
\item \textbf{Diagnosis.} Through controlled backend replacements, we identify planning as the dominant bottleneck of omni-modal agents.
\item \textbf{System.} Evidence-ledger planning keeps observation noise out of persistent context: each task maintains a typed ledger, the critic digests every observation into ledger updates, the ledger accepts updates only through the critic's verdicts, and the planner's persistent context contains only the ledger, with each raw observation seen once at the step under review.
\item \textbf{Training.} A recipe that requires no step-level manual annotation: the trajectories and ledger states recorded during execution are harvested as two levels of training signal, supervised fine-tuning and decision-level reinforcement learning, which improve two planner backbones.
\end{itemize}
Omni-Decision reaches 81.4\% accuracy on open-world tool interaction in OmniGAIA and 65.0\% on long-video understanding in WorldSense, at a cost per question well below that of direct answering by the strongest model, Gemini-3.1-Pro.

\section{Related Work}
\label{sec:related-work}

\paragraph{Omni-modal agent systems.}
Existing systems advance long-horizon multimodal QA through multi-role orchestration, planner--critic collaboration, tool interfaces, and evidence chain modeling \citep{kumarMMCTAgentMultimodalCritical2024,taoOmniAgentAudioGuidedActive2025,luVITEDVideoTemporal,liuLongVideoAgentMultiAgentReasoning2025,liuVideoMindChainofLoRAAgent2025,yangLongVTIncentivizingThinking2025,zhuOmniRAGAgentAgenticOmnimodal2026}. Orchestra-o1 decomposes tasks by modality, runs specialized sub-agents in parallel, and post-trains the orchestrator with decision-level RL \citep{zhangOrchestraO1Omnimodal2026}, while sandboxed coding agents externalize perception and extract evidence from media as environment state \citep{chenSandboxedCodingAgents2026}. OmniAgent writes evidence acquired on demand into textual memory and combines this memory with agentic post-training \citep{xingNativeActivePerception2026}. These systems carry intermediate observations in free-form traces, file workspaces, or untyped textual memory, and planning operates directly on those sequences. Omni-Decision uses a task-scoped evidence ledger as its central control object: each observation is digested into a typed evidence event before admission, and planning, verification, and stopping all read the same ledger.

\paragraph{Verification and explicit state.}
Verification checks and judgments of answer readiness improve reasoning reliability \citep{shinnReflexionLanguageAgents2023,hanVerifiAgentUnifiedVerification,maDoVerInterventionDrivenAuto2026}, and agent memory and step-level process trajectories externalize intermediate information in several forms \citep{packerMemGPTTowardsLLMs2023,zhuWhereLLMAgents2025a,agentPRMStepLevelProcessReward2026}. Omni-Decision differs in where verified information is stored: the reducer commits critic verdicts as typed events that condition every later planning step. The ledger retains only the evidence needed to close the current query; its scope does not extend to open-ended organization of long-term memory.

\paragraph{Context engineering.}
ACON learns to compress observations and interaction histories for long-horizon agents \citep{kangACONOptimizingContext2025}; the Complexity Trap study finds that replacing stale observations with placeholders can match LLM summarization \citep{lindenbauerComplexityTrap2025}. These methods compress unstructured messages after they enter history. Omni-Decision keeps persistent context clean by construction: the critic digests each observation into typed state updates on arrival, and the planner's persistent context contains only control signals throughout the task.

\begin{figure}[t]
  \centering
  \includegraphics[width=0.9\linewidth]{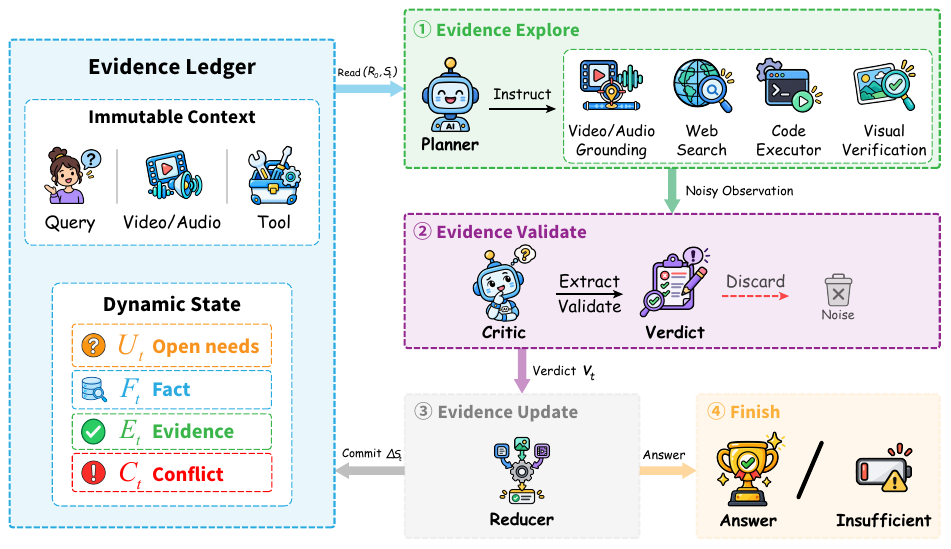}
  \caption{The Omni-Decision control loop. At each step, the planner selects an action from the current ledger, the critic checks the returned observation, and the reducer commits a typed event to the ledger.}
  \label{fig:omnidecision-state-dataflow}
\end{figure}

\section{Method: Omni-Decision}
\label{sec:method}

This section first presents an overview of evidence-ledger planning (Section~\ref{sec:method-overview}), then defines the ledger state (Section~\ref{sec:method-ledger}) and the state-conditioned control and reduction rules (Section~\ref{sec:method-control}), and finally describes the training recipe (Section~\ref{sec:method-training}).

\subsection{Overview}
\label{sec:method-overview}

For each task, Omni-Decision maintains an evidence ledger, a typed evidence state that records open evidence needs, fact and computation dependencies, confirmed evidence atoms, and unresolved conflicts. Multimodal observations such as web pages, video frames, and subtitles contain substantial noise. In a transient call, the critic digests each new observation against the ledger and produces a typed evidence event. The planner's persistent dialogue receives only the ledger's control projection (open needs, gap diagnoses, and readiness signals), together with the raw observation currently under review. Noise therefore does not accumulate in any persistent context. Planning proceeds over a clean context whose persistent portion stays approximately constant in length across steps. Once all blocking needs are closed and no conflicts remain, the ledger declares readiness and the system drafts an answer against it.

Figure~\ref{fig:omnidecision-main-flow} traces this mechanism on an OmniGAIA task: four evidence needs parsed from the query are closed through video grounding, web search, and computation, after which the ledger is ready.

\subsection{Task-scoped evidence ledger}
\label{sec:method-ledger}

Before any tool call, Omni-Decision initializes the evidence ledger from the query: the planner parses the question into a checklist of open evidence needs---in Figure~\ref{fig:omnidecision-main-flow}, the brand's identity, the required dates, and the month difference---together with the fact and computation slots these needs must fill. At step $t$ the ledger state is $S_t=\{U_t,F_t,E_t,C_t\}$: $U_t$ holds the unresolved evidence needs (the open needs), $F_t$ tracks fact slots, entity attributes, and computation results, $E_t$ stores confirmed evidence atoms with source and temporal support, and $C_t$ records contradictions between observations that still await reconciliation; $E$ and $C$ start empty. As observations are committed, needs leave $U_t$ while verified values fill $F_t$ and the supporting atoms enter $E_t$, and the task can be answered once every blocking need on the checklist is closed. Each field supports a distinct control decision: $U_t$ guides the next action to collect evidence, empty slots in $F_t$ identify missing facts and computations, $E_t$ provides sourced support for the final answer, and any unresolved conflict in $C_t$ blocks an answer.

The ledger is task-scoped, and it separates fixed task inputs from this stepwise state (Figure~\ref{fig:omnidecision-state-dataflow}, left): the original query, pointers to multimodal assets, the answer format, and the available modalities and tools remain unchanged as immutable context $R_0$. The planner therefore receives the immutable task context $R_0$ and the current ledger state $S_t$, rendered as bounded text, together with the observation returned at the previous step, which is shown once and not retained. Appendix~\ref{sec:appendix-ledger} gives the complete ledger form, including the entry structure of each field and two fully instantiated execution traces.

\subsection{State-conditioned control and reduction}
\label{sec:method-control}

At each step, the planner reads the current ledger. The runtime renders $S_t$ as bounded text that reports the state of the evidence checklist: which needs remain open, which fact slots have been filled, which evidence has been confirmed, which conflicts remain unresolved, and what is still missing before an answer can be produced. From the available action set $\mathcal{A}(R_0)$, which is determined by the assets, tools, and answer format, the planner selects the next action $a_t$. It may invoke media grounding, retrieval, browsing, computation, or visual verification, or it may select \texttt{finish}.

After a tool returns observation $o_t$, the observation does not write to the ledger directly. In a transient call, the critic digests $o_t$ against the current evidence needs, determines whether it supports, conflicts with, or omits the required evidence, and produces verdict $v_t$. Separating verification from planning subjects each execution step to two independent judgments: the critic checks whether the raw observation is sufficient, and the planner chooses the next action from the verified ledger. If the planner performed both tasks, it would have to evaluate its newly obtained observation while making the next decision. LLMs do not reliably correct their own reasoning errors without external feedback \citep{huangLargeLanguageModelsCannot2024}, so an error that treats a noisy page as a closed need can persist throughout the dialogue history. An independent critic intercepts such an error at the commit boundary and marks the evidence insufficient before it enters the ledger.

The reducer is the ledger's sole writer. It converts accepted observations or verdicts into typed events and commits the state increment
\[
S_{t+1}=S_t\oplus\Delta S_t.
\]
The operator $\oplus$ applies deterministic updates field by field. It removes satisfied needs from $U_t$, fills verified facts and computation results in $F_t$, appends new evidence atoms to $E_t$, and records events that contradict existing evidence in $C_t$ while retaining both sources. In Figure~\ref{fig:omnidecision-main-flow}, one such commit fills the launch date slot in $F_t$, appends the sourced atom to $E_t$, and removes the corresponding need from $U_t$.

Tool and LLM outputs may be stochastic, but the ledger update rules are deterministic. Randomness stops at the commit boundary, and no module can bypass the reducer to modify the ledger directly: all state changes pass through one controlled channel, and all dependencies are read as explicitly declared, read-only inputs \citep{shiProgrammingParadigmSpatiotemporal2026}.

Readiness follows directly from the ledger:
\[
\mathrm{ready}(S_t)=(U^{\mathrm{b}}_t=\varnothing)\wedge\mathrm{complete}(F_t)\wedge(C_t=\varnothing).
\]
When parsing the query, the planner marks each need as blocking or supplementary, and $U^{\mathrm{b}}_t\subseteq U_t$ denotes the open blocking needs. The three conjuncts state the conditions that must hold before answering: no blocking need remains open, every fact and computation slot implied by the query holds a verified value, and no conflict remains unresolved. A contradiction between observations stays in $C_t$ and blocks an answer until new evidence resolves it. A supplementary need may stay open once the answer no longer depends on it, and a need that repeated attempts cannot close is marked unfillable and leaves $U_t$ as terminal; a correct run can therefore end with some declared needs still open, as the audit in Section~\ref{sec:experiments-progress} records. When the planner selects \texttt{finish}, the finalizer drafts a candidate answer and checks it against the ledger. If the check fails, the reducer writes the diagnosis of missing evidence or conflict back to the ledger as an event for the next planning step.

The ledger also determines when the system should stop. Before selecting an action, the system checks whether any worthwhile action remains. Such an action must be available under $R_0$ and the remaining budget, address an open need, empty slot, or unresolved conflict in the ledger, and not be exhausted by repeated failures. If the ledger is not ready and no action meets these conditions, the system terminates and reports insufficient evidence. If repeated \texttt{finish} attempts exhaust the revision budget, the system returns its current best value, marked as forced (Appendix~\ref{sec:appendix-failure}).

\begin{algorithm}[t]
\caption{Evidence-ledger inference loop}
\label{alg:omni-decision}
\small\begin{algorithmic}[1]
\Require query $q$, immutable context $R_0$, action space $\mathcal{A}(R_0)$, maximum steps $T$
\State initialize ledger $S_0=\{U_0,F_0,E_0,C_0\}$ from $q$ and assets
\For{$t=0,\ldots,T-1$}
  \State $a_t\gets\mathrm{decide}(R_0,S_t,o_{t-1};\mathcal{A}(R_0))$ \Comment{planner reads the rendered ledger and the latest observation; $o_{-1}=\varnothing$}
  \If{$a_t=\texttt{finish}$}
    \State $\hat{y}\gets\mathrm{draft}(R_0,S_t)$
    \If{$\mathrm{check}(\hat{y},S_t)$ passes} \State \Return $\hat{y}$
    \Else \State $S_{t+1}\gets\mathrm{reduce}(S_t,\text{blocked diagnosis})$
    \EndIf
  \Else
    \State $o_t\gets\mathrm{execute}(a_t)$;\; $v_t\gets\mathrm{critic}(o_t,S_t)$
    \State $S_{t+1}\gets\mathrm{reduce}(S_t,v_t)$ \Comment{normalization, need matching, validation: Appendix~\ref{sec:appendix-implementation}}
  \EndIf
  \If{$\neg\mathrm{ready}(S_{t+1})$ and no remaining action can advance the ledger}
    \State \Return \texttt{insufficient}
  \EndIf
\EndFor
\State \Return \texttt{insufficient}
\end{algorithmic}
\end{algorithm}

Algorithm~\ref{alg:omni-decision} summarizes the inference loop: the planner reads the ledger to select an action, the critic digests the observation, and the reducer records the resulting event. The ledger determines both readiness and stopping. Appendix~\ref{sec:appendix-implementation} gives the implementation semantics for observation normalization, need matching, and evidence-level validation.

\subsection{Training recipe}
\label{sec:method-training}

The ledger mechanism requires no training at inference time, but its execution traces provide process supervision without step-level annotation: every planner decision is recorded together with the ledger state it saw and the critic verdict that followed.

We use these traces in two stages. State-SFT retains runs that the judge marks as correct, expands each step into a pair of ledger state and planner action, and fine-tunes the planner role. Closure-aligned reinforcement learning then targets the dominant failure of weak planners, abstaining or converging while needs remain open (Section~\ref{sec:experiments-backend}): for each ledger state reconstructed from a correct trace, the policy samples a group of candidate decisions and reinforces those scored above the group mean by rule-based closure checks and a lightweight reward judge. Because the ledger lists each open need explicitly, a single decision can be scored without executing it, so this stage requires no system rollout. Appendix~\ref{sec:appendix-training} gives the data construction, reward definitions, and optimization objective; Section~\ref{sec:experiments-training} reports the results.

\section{Experiments}
\label{sec:experiments}

\subsection{Experimental setup}
\label{sec:experiments-setup}

\paragraph{Benchmarks.} OmniGAIA is the primary benchmark, containing 360 open-world evidence-seeking questions spanning video, audio, images, web evidence, and computation \citep{liOmniGAIANativeOmniModal2026}. WorldSense is a complementary conventional long-video benchmark: answers are contained in video, audio, and subtitles, questions are multiple-choice, and web search is disabled \citep{hongWorldSenseEvaluatingRealworld2025}.

\paragraph{Implementation details.} The planner, critic, and finalizer are instantiated by one LLM that reads only the ledger and the current observation as text; media never enter their input. Perception is exposed to the planner as a tool, and the planner invokes it according to the open needs in the ledger. The main experiments use GPT-5.2 for these roles and Gemini-3.1-Pro for perception; experiments that replace a backend state the replacement in place. Exact model versions, tool configuration, and prompt templates appear in Appendix~\ref{sec:appendix-implementation} and Appendix~\ref{sec:appendix-prompts}.

\subsection{OmniGAIA results}
\label{sec:experiments-main}

\begin{table*}[t]
\caption{OmniGAIA results. $^\ast$ denotes publicly reported results from the official leaderboard \citep{liOmniGAIANativeOmniModal2026} or original papers; $^\dagger$ denotes our measurements.}
\label{tab:omnigaia-main}
\centering
\scriptsize
\setlength{\tabcolsep}{3pt}
\renewcommand{\arraystretch}{1.2}
\resizebox{\textwidth}{!}{%
\begin{tabular}{lr@{\hspace{8pt}}rrr@{\hspace{8pt}}rrrrrrrrr}
\toprule
\multirow{2}{*}{\textbf{Method}} & \multirow{2}{*}{\textbf{Overall}} & \multicolumn{3}{@{}c@{\hspace{8pt}}}{\textbf{Difficulty}} & \multicolumn{9}{c}{\textbf{Category}}\\
\cmidrule(l{0.2em}r{1em}){3-5}\cmidrule(lr){6-14}
 & & Easy & Medium & Hard & Geo. & Tech. & Hist. & Fin. & Sports & Art & Movie & Sci. & Food \\
\midrule
\rowcolor{gray!15}\multicolumn{14}{c}{\textit{End-to-end models}}\\
Qwen3-Omni-30B$^\ast$ & 13.30&19.70&10.60&9.00&8.70&14.30&11.90&28.00&10.80&13.90&9.10&15.40&22.20\\
Qwen3.5-Omni-Flash$^\ast$ & 33.90&--&--&--&--&--&--&--&--&--&--&--&--\\
Qwen3.5-Omni-Plus$^\ast$ & 57.20&--&--&--&--&--&--&--&--&--&--&--&--\\
Gemini-2.5-Pro$^\ast$ &30.80&41.80&26.90&21.80&23.20&28.60&32.80&20.00&32.40&41.70&42.40&26.90&33.30\\
Gemini-3-Flash$^\ast$ &51.70&67.20&46.90&37.20&50.70&57.10&44.80&48.00&59.50&55.60&54.60&38.50&61.10\\
Gemini-3-Pro$^\ast$ &62.50&78.70&61.90&38.50&65.20&59.20&62.10&72.00&78.40&52.80&48.50&42.30&88.90\\
Gemini-3.1-Pro$^\dagger$ &79.44&86.07&79.38&69.23&85.51&79.59&80.30&84.00&83.78&69.44&63.64&80.77&83.33\\
\midrule
\rowcolor{gray!15}\multicolumn{14}{c}{\textit{Agent systems}}\\
Minimal agent$^\dagger$ &5.56&9.02&3.12&5.13&1.45&10.20&10.61&0.00&2.70&2.78&6.06&3.85&11.11\\
OmniGAIA base$^\dagger$ &18.33&26.23&17.50&7.69&5.80&28.57&28.79&16.00&13.51&13.89&15.15&26.92&11.11\\
OmniAtlas-Qwen3-30B$^\ast$ &20.80&31.10&18.80&9.00&10.10&30.60&29.90&32.00&18.90&16.70&12.10&11.50&27.80\\
OmniAgent$^\dagger$ &25.51&31.59&24.59&17.89&20.30&28.58&28.86&22.41&16.65&29.57&30.56&26.89&28.01\\
Orchestra-o1-GPT-5$^\ast$ &72.80&80.30&75.00&56.40&72.50&69.40&75.80&64.00&83.80&63.90&69.70&73.10&83.30\\
Sandboxed agent$^\ast$ &75.00&82.00&75.00&64.10&--&--&--&--&--&--&--&--&--\\
\textbf{Omni-Decision (ours)}$^\dagger$ &\textbf{81.39}&90.98&78.75&71.79&82.61&85.71&81.82&80.00&86.49&72.22&78.79&80.77&77.78\\
\bottomrule
\end{tabular}}
\end{table*}

Omni-Decision reaches 81.39\% overall accuracy on OmniGAIA, the best result on the benchmark (Table~\ref{tab:omnigaia-main}). Gemini-3.1-Pro, measured in the same evaluation window with the same judge, reaches 79.44\%; Omni-Decision is 1.95 points higher, with the difference concentrated on Easy and Hard. The strongest publicly reported agent system, the sandboxed coding agent \citep{chenSandboxedCodingAgents2026}, reaches 75.00\%; Omni-Decision is 6.4 points higher overall and 7.7 points higher on Hard, the largest margin among the three difficulty levels. The other three agent systems marked $^\dagger$ were run by us in the same window with the same pair of backends: Minimal agent gives the model tool access without any control structure for organizing tool outputs, OmniGAIA base is the official baseline pipeline, and OmniAgent is the audio-guided active agent of \citet{taoOmniAgentAudioGuidedActive2025}. All three stay below 26\%. With the same pair of backends, overall accuracy ranges from 5.56\% to 81.39\% across harnesses, so the harness decides how much of the models' capability is realized, in line with the observation cited in Section~\ref{sec:introduction} \citep{zhangStopComparingAgents2026}. Cost is computed from billing records: Omni-Decision averages approximately \$1.2 per question, and Gemini-3.1-Pro approximately \$2.8 in the same window, so Omni-Decision achieves higher accuracy at roughly 43\% of the cost. Appendix~\ref{sec:appendix-cost} reports the cost distribution across questions.

\subsection{Backend sensitivity}
\label{sec:experiments-backend}
\label{sec:experiments-ablation}

\begin{table}[t]
\caption{Controlled backend replacements. Each row replaces only the indicated backend; all other components, prompts, and budgets match the default configuration in the first row.}
\label{tab:backend-sensitivity}
\centering
\small
\begin{tabular}{ll@{\hspace{14pt}}rrr@{\hspace{14pt}}r}
\toprule
\multirow{2}{*}{\textbf{Planner}} & \multirow{2}{*}{\textbf{Perception}} & \multicolumn{3}{@{}c@{\hspace{14pt}}}{\textbf{Difficulty}} & \multirow{2}{*}{\textbf{Overall}}\\
\cmidrule(l{0.2em}r{1.45em}){3-5}
 & & Easy & Medium & Hard & \\
\midrule
GPT-5.2 & Gemini-3.1-Pro &90.98&78.75&71.79&\textbf{81.39}\\
\midrule
\rowcolor{gray!15}\multicolumn{6}{c}{\textit{Replace planner}}\\
GLM-5.2 & Gemini-3.1-Pro &84.43&71.25&67.95&75.00\\
Qwen3.5-Plus & Gemini-3.1-Pro &72.95&51.25&48.72&58.06\\
Qwen3.5-27B & Gemini-3.1-Pro &62.30&41.25&34.62&46.94\\
Qwen3-Omni-30B & Gemini-3.1-Pro &21.31&11.25&11.54&14.72\\
\midrule
\rowcolor{gray!15}\multicolumn{6}{c}{\textit{Replace perception}}\\
GPT-5.2 & Qwen3.5-Omni-Plus &74.59&64.38&53.85&65.56\\
GPT-5.2 & Qwen3.5-Omni-Flash &68.03&57.50&47.44&58.89\\
GPT-5.2 & Qwen3-Omni-30B &72.13&63.13&52.56&63.89\\
\midrule
\rowcolor{gray!15}\multicolumn{6}{c}{\textit{Replace both}}\\
Qwen3-Omni-30B & Qwen3-Omni-30B &20.49&10.63&10.26&13.89\\
\bottomrule
\end{tabular}
\end{table}

We hold the rest of the system fixed and replace one backend at a time (Table~\ref{tab:backend-sensitivity}). Replacing the planner affects accuracy far more than replacing perception. With Gemini-3.1-Pro perception fixed, moving the planner from GPT-5.2 through GLM-5.2, Qwen3.5-Plus, and Qwen3.5-27B lowers overall accuracy step by step from 81.39\% to 46.94\%, and Qwen3-Omni-30B leaves only 14.72\%. With the GPT-5.2 planner fixed, all three perception replacements stay at or above 58.89\%. The Plus tier of the Qwen3.5 family gives the most direct comparison within one model generation: replacing the planner costs 23.3 points and replacing perception costs 15.8 points, a ratio of approximately 1.5. This is direct evidence for the claim in Section~\ref{sec:introduction}: on these tasks the main shortfall of current multimodal models lies in planning, and the gap on the perception side is much smaller. Once the planner is weak enough, perception quality no longer matters: replacing both backends gives 13.89\%, nearly the same as the 14.72\% from replacing only the planner. The failures of the weak planner are predominantly decision-level behaviors, namely abstention or convergence before the evidence is closed.

\subsection{State ablation}
\label{sec:experiments-state}

\begin{table}[t]
\caption{State ablation. All three configurations share the default backends, tools, and step limit.}
\label{tab:state-ablation}
\centering
\begin{tabular}{l@{\hspace{14pt}}rrr@{\hspace{14pt}}r}
\toprule
\multirow{2}{*}{\textbf{Configuration}} & \multicolumn{3}{@{}c@{\hspace{14pt}}}{\textbf{Difficulty}} & \multirow{2}{*}{\textbf{Overall}}\\
\cmidrule(l{0.2em}r{1.3em}){2-4}
 & Easy & Medium & Hard & \\
\midrule
Evidence Ledger &90.98&78.75&71.79&\textbf{81.39}\\
Memory &77.05&63.13&65.38&68.33\\
No-ledger ReAct &74.59&55.00&48.72&60.28\\
\bottomrule
\end{tabular}
\end{table}

We use two controls to isolate the ledger's contribution. No-ledger ReAct removes the ledger entirely: the planner reads the raw observations accumulated chronologically in the dialogue history, as in standard ReAct. Memory follows the common design of agent memory components: successive observations are compressed into a rolling natural language summary, and the planner reads this summary at each step instead of the raw history; Appendix~\ref{sec:appendix-implementation} gives implementation details. Both controls expose prior observations to the planner and therefore achieve nontrivial accuracy, but the full method leads both on every split. Relative to No-ledger ReAct, the gain is approximately 24 points on Medium and Hard, compared with approximately 16 points on Easy. Longer tasks place greater demands on tracking which needs are closed and which remain open, a distinction that raw history and natural language summaries do not maintain reliably as tasks lengthen. Per-call token usage of the three configurations is reported in Appendix~\ref{sec:appendix-cost}.

\subsection{Training}
\label{sec:experiments-training}

\begin{table}[t]
\caption{Training on ledger trajectories. Evaluation fixes Gemini-3.1-Pro perception. The base row of each planner is the untrained backbone and matches the corresponding row in Table~\ref{tab:backend-sensitivity}. Easy, Medium, and Hard use the complete subsets of 122, 160, and 78 questions.}
\label{tab:ledger-training}
\centering
\begin{tabular}{l@{\hspace{14pt}}rrr@{\hspace{14pt}}r}
\toprule
\multirow{2}{*}{\textbf{Planner}} & \multicolumn{3}{@{}c@{\hspace{14pt}}}{\textbf{Difficulty}} & \multirow{2}{*}{\textbf{Overall}}\\
\cmidrule(l{0.2em}r{1.3em}){2-4}
 & Easy & Medium & Hard & \\
\midrule
Qwen3.5-27B &62.30&41.25&34.62&46.94\\
\quad+ state-SFT &65.57&43.75&35.90&49.44\\
\quad\quad+ RL &66.39&45.00&37.18&\textbf{50.56}\\
\midrule
Qwen3-Omni-30B &21.31&11.25&11.54&14.72\\
\quad+ state-SFT &25.41&15.00&15.38&\textbf{18.61}\\
\bottomrule
\end{tabular}
\end{table}

We test whether ledger trajectories can train weaker planners (Table~\ref{tab:ledger-training}; Section~\ref{sec:method-training} gives the recipe and Appendix~\ref{sec:appendix-training} the details); each base model and its trained counterpart share an otherwise identical evaluation configuration and can be compared directly. Starting from Qwen3.5-27B, state-SFT followed by closure-aligned RL raises overall accuracy from 46.94\% to 50.56\%, with consistent gains on Medium and Hard, and state-SFT alone improves Qwen3-Omni-30B in the same direction, from 14.72\% to 18.61\%. Training gains vary with the planner backbone's initial capability, consistent with the gradient in Section~\ref{sec:experiments-backend}.

\subsection{Evidence progress audit}
\label{sec:experiments-progress}

\begin{figure}[t]
  \centering
  \includegraphics[width=\linewidth]{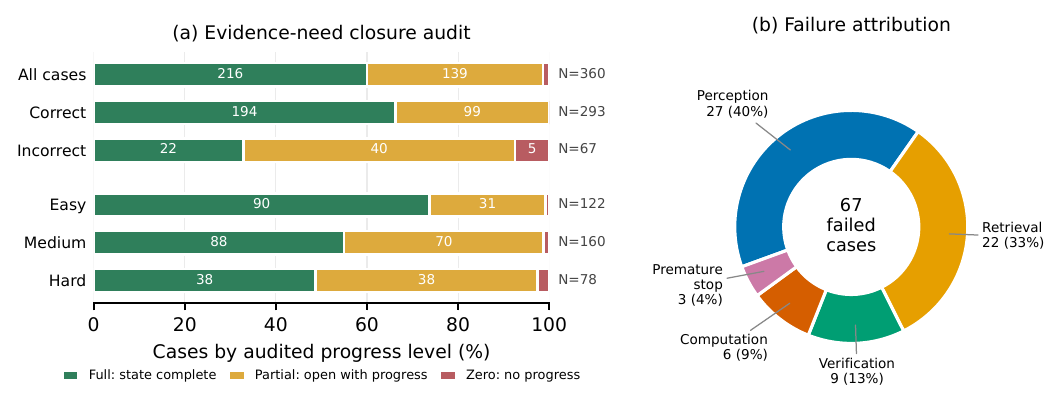}
  \caption{Evidence progress audit. (a) Evidence-need closure across all 360 questions. (b) Attribution of the first decisive cause in the 67 failed cases.}
  \label{fig:progress-audit}
\end{figure}

The ledger records the evidence needs declared and closed for each task, which allows us to audit evidence progress across all 360 questions (Figure~\ref{fig:progress-audit}(a)). Correct cases close about twice the share of their declared needs as incorrect cases, and most incorrect cases still close at least one need: the ledger sustains evidence progress through long tasks, and most failures occur when a single link of the evidence chain remains open. Attributing the first decisive cause of this last unresolved link (Figure~\ref{fig:progress-audit}(b)) concentrates residual failures in evidence acquisition and verification: media perception and external retrieval together account for approximately three quarters, and premature stopping at the decision layer accounts for only a small fraction. This agrees with Section~\ref{sec:experiments-backend}: with the ledger and strongest planner, the decision layer is no longer the main failure source, and evidence acquisition becomes the remaining bottleneck. Appendix~\ref{sec:appendix-failure} gives the exact counts and per-case analysis.

\subsection{WorldSense}
\label{sec:experiments-worldsense}

\begin{table*}[t]
\caption{WorldSense results. Avg.\ is accuracy over all questions; $^\ast$ denotes public leaderboard results \citep{hongWorldSenseEvaluatingRealworld2025}, $^\dagger$ denotes our measurements.}
\label{tab:worldsense}
\centering
\small
\setlength{\tabcolsep}{3.5pt}
\renewcommand{\arraystretch}{1.15}
\begin{tabular}{lr@{\hspace{9pt}}rrrrrrrr}
\toprule
\multirow{2}{*}{\textbf{Method}} & \multirow{2}{*}{\textbf{Avg.}} & \multicolumn{8}{c}{\textbf{Domain}}\\
\cmidrule(lr){3-10}
 & & Tech. & Cult. & Daily & Film & Perf. & Games & Sports & Music\\
\midrule
\rowcolor{gray!15}\multicolumn{10}{c}{\textit{End-to-end models}}\\
Qwen3-Omni$^\ast$ &54.00&58.70&60.50&54.50&53.80&55.40&46.80&48.80&52.20\\
video-SALMONN 2+$^\ast$ &56.50&59.00&63.10&54.00&59.90&58.10&54.10&51.90&54.40\\
Qwen3.5-Omni-Flash$^\ast$ &57.80&59.00&59.90&57.30&60.20&59.90&52.40&54.20&58.40\\
Qwen3.5-Omni-Plus$^\ast$ &62.80&66.90&66.00&62.20&65.20&68.90&57.10&56.30&61.60\\
Gemini-2.5-Pro$^\ast$ &65.10&64.90&66.00&65.80&68.10&69.70&65.70&63.50&61.30\\
Gemini-3.1-Pro$^\ast$ &\textbf{65.50}&67.30&66.70&67.50&68.60&70.80&61.80&59.30&63.30\\
\midrule
\rowcolor{gray!15}\multicolumn{10}{c}{\textit{Agent systems}}\\
OmniAgent$^\dagger$ &28.06&33.27&20.06&14.29&58.31&12.36&42.92&14.19&38.42\\
\textbf{Omni-Decision (ours)}$^\dagger$ &65.00&64.94&77.08&58.65&71.67&71.43&62.16&55.88&67.19\\
\bottomrule
\end{tabular}
\end{table*}

WorldSense tests the same system on conventional long-video understanding. Answers are contained in video, audio, and subtitles; web search is disabled, so all evidence comes from media tools. Omni-Decision obtains 65.0\% (Table~\ref{tab:worldsense}), on par with the leading end-to-end models on the public leaderboard, including Gemini-3.1-Pro at 65.5\%, and higher than the other models and agent systems in the table. It averages approximately \$0.3 per question, well below the approximately \$0.8 cost of direct answering by the same model (Appendix~\ref{sec:appendix-cost}).

\section{Discussion}
\label{sec:discussion}

Omni-Decision makes the system layer an explicit object, and this is what allows the bottlenecks to be attributed. Externally acquired evidence enters planning only as typed state, so the contribution of this layer can be defined and ablated on its own, and the planner and perception backends can be replaced one at a time with everything else held fixed. The controlled replacements give a clear order (Section~\ref{sec:experiments-backend}): on these tasks the main shortfall of current multimodal models lies in planning, and once the planner is weak enough, perception quality has almost no effect on the outcome. The same ledger locates the remaining errors one layer down: with the ledger and the strongest planner in place, failures concentrate in media perception and external retrieval (Section~\ref{sec:experiments-progress}). Together, planning is the shortfall to close first, and perception is the bottleneck that appears once planning is closed.

Post-training on ledger trajectories improves both Qwen planner backbones (Section~\ref{sec:experiments-training}). The current design has one scope boundary: the critic reads observations only in text form, and media never enter its input, so ledger quality is bounded by the text output of the perception tools. A critic that reads multimodal input directly is the natural next step.

\section{Conclusion}
\label{sec:conclusion}

We present Omni-Decision to address the planning bottleneck of omni-modal agents. Each task maintains a typed evidence ledger; the critic processes each observation and keeps only the usable evidence, so the planner decides throughout on a compact, verified state. Omni-Decision reaches state-of-the-art accuracy of 81.4\% on open-world tool interaction in OmniGAIA at approximately 43\% of Gemini-3.1-Pro's inference cost, and 65.0\% on long-video understanding in WorldSense. The ledger states, actions, and verdicts recorded during execution form a training signal that requires no step-level human annotation, and supervised fine-tuning with decision-level reinforcement learning on this signal consistently improves two weaker planner backbones. The system produces trajectories, and the trajectories in turn train the planner.

\bibliography{OmniDecision}
\bibliographystyle{plainnat}

\clearpage
\raggedbottom
\appendix
\numberwithin{table}{section}
\numberwithin{figure}{section}

\section{Ledger Anatomy and Execution Traces}
\label{sec:appendix-ledger}

This appendix presents the complete ledger form. Table~\ref{tab:appendix-ledger-fields} summarizes what each of its four fields records, the form of its entries, and the control decision it drives. We then give two fully instantiated trajectories from actual run logs and map their variables to the mechanism in Figure~\ref{fig:omnidecision-main-flow}. Section~\ref{sec:appendix-trace-158} shows normal evidence closure, while Section~\ref{sec:appendix-trace-323} shows how a conflict is recorded and resolved. Neither run receives a reference answer as input.

\begin{table}[htbp]
\caption{Ledger fields, entry forms, and control roles. Examples are taken from the trajectories in Sections~\ref{sec:appendix-trace-158} and~\ref{sec:appendix-trace-323}.}
\label{tab:appendix-ledger-fields}
\centering\scriptsize
\setlength{\tabcolsep}{4pt}
\begin{tabular}{@{}p{0.14\linewidth}p{0.28\linewidth}p{0.34\linewidth}p{0.18\linewidth}@{}}
\toprule
Field & Recorded content & Entry form & Control role\\
\midrule
$U_t$: open evidence needs & Evidence-collection checklist; each need has a target description and closure status. & ``Anchor the Fiji scene date'' (\texttt{unfilled}). & Selects the next action to collect evidence.\\
$F_t$: fact and computation slots & Query-implied external facts, entity attributes, and computation results. & ``Date difference = 81 days'' (verified). & Marks missing facts and computations.\\
$E_t$: confirmed evidence & Evidence atoms with source and temporal support. & Title card ``Fiji Islands, South Pacific, Oct 26th'' (00:02:18--00:02:31, \texttt{frame\_confirm}). & Supports the final answer.\\
$C_t$: unresolved conflicts & Contradictions between observations, with type, severity, and linked evidence. & \texttt{temporal\_mismatch}: the end card gives the date only to the month, while external retrieval gives an exact date. & Blocks finalization until resolved.\\
\bottomrule
\end{tabular}
\end{table}

\subsection{OmniGAIA \#158: normal closure}
\label{sec:appendix-trace-158}

The question asks for the date difference between Fiji and Los Angeles in an 18:10 travel video, the corresponding number of complete weeks and remaining days, and the effect of a leap day. Table~\ref{tab:appendix-158-initial} gives the initial ledger, Table~\ref{tab:appendix-158-trace} the step-by-step trace, and Table~\ref{tab:appendix-158-terminal} the terminal ledger.

\begin{table}[htbp]
\caption{Initial ledger for OmniGAIA \#158.}
\label{tab:appendix-158-initial}
\centering\small
\begin{tabular}{p{0.13\linewidth}p{0.78\linewidth}}
\toprule Component & Content\\ \midrule
$U_0$ & Needs $u_1$--$u_5$: Fiji date, Los Angeles date, date difference, complete weeks and remaining days, and leap-day effect; all are \texttt{unfilled}.\\
$F_0$ & Pending slots for the date difference, complete weeks and remaining days, and the leap-day effect.\\
$E_0$ & Empty. The query provides only three cues: the Fiji scene is near 150 s, the Los Angeles scene is near 680 s, and narrative time moves forward.\\
$C_0$ & Empty.\\
\bottomrule
\end{tabular}
\end{table}

\begin{table}[htbp]
\caption{Step-by-step trace for OmniGAIA \#158.}
\label{tab:appendix-158-trace}
\centering\scriptsize
\setlength{\tabcolsep}{3pt}
\begin{tabular}{@{}p{0.06\linewidth}p{0.32\linewidth}p{0.25\linewidth}p{0.12\linewidth}p{0.14\linewidth}@{}}
\toprule Step & Action and observation & Ledger update & Critic verdict & Ready to finalize?\\ \midrule
$t=1$ & Locate candidate intervals: Fiji 00:02:20--00:02:29 and Los Angeles 00:11:10--00:11:19. & $u_1,u_2\to\texttt{partial}$; next narrow the intervals to confirm the title cards. & INSUFFICIENT & No; all five needs remain open.\\
$t=2$ & 00:02:18--00:02:31 shows ``Fiji Islands, South Pacific, Oct 26th.'' & $u_1\to$ filled; title card and interval enter $E$. & INSUFFICIENT & No; $u_2$--$u_5$ remain.\\
$t=3$ & 00:11:15--00:11:21 shows ``Los Angeles, January 15th'' without a year. & $u_2\to$ filled; evidence enters $E$. & INSUFFICIENT & No; $u_3$--$u_5$ remain.\\
$t=4$ & Narrative order gives 81 days from Oct. 26 to Jan. 15 of the next year; $81=11\times7+4$; the interval does not include Feb. 29. & $u_3$--$u_5\to$ filled; all three slots receive their values. & SUFFICIENT & Yes.\\
$t=5$ & Final check and output. & Finalization only. & PASS & Submit.\\
\bottomrule
\end{tabular}
\end{table}

\begin{table}[htbp]
\caption{Terminal ledger for OmniGAIA \#158 at finalization step $t=5$.}
\label{tab:appendix-158-terminal}
\centering\small
\begin{tabular}{@{}p{0.12\linewidth}p{0.52\linewidth}p{0.27\linewidth}@{}}
\toprule Component & Content & Readiness condition\\
\midrule
$U$ & Needs $u_1$--$u_5$ are all filled. & $U_t=\varnothing$: satisfied.\\
$E$ & Two title cards and their intervals, 00:02:18--00:02:31 and 00:11:15--00:11:21. & Not applicable.\\
$C$ & No conflict was recorded during the trajectory. & $C_t=\varnothing$: satisfied.\\
$F$ & 81 days; 11 weeks and 4 days; leap-day adjustment 0. & $\mathrm{complete}(F_t)$: satisfied.\\
\bottomrule
\end{tabular}
\end{table}

All three conjuncts hold, so $\mathrm{ready}(S_T)=\mathrm{true}$. The system submits ``81 days, or 11 complete weeks and 4 days; the leap-day increment is 0,'' which matches the reference answer.

\subsection{OmniGAIA \#323: conflict recording and resolution}
\label{sec:appendix-trace-323}

The task provides behind-the-scenes footage from a Harry Potter film and a news segment about a Harry Potter-themed bar in Toronto. It asks for the exact age difference, on the film's premiere date, between the actor who played the bar's namesake character and the actress in the behind-the-scenes footage. Each $C_t$ entry records its type, severity, description, and linked evidence. Any unresolved entry blocks finalization. Table~\ref{tab:appendix-323-conflict} records how the conflict is written into the ledger and resolved.

\begin{table}[htbp]
\caption{Conflict trace for OmniGAIA \#323.}
\label{tab:appendix-323-conflict}
\centering\small
\setlength{\tabcolsep}{3pt}
\begin{tabular}{@{}p{0.10\linewidth}p{0.57\linewidth}rp{0.16\linewidth}@{}}
\toprule Step & Runtime event & $|C_t|$ & Ready to finalize?\\ \midrule
$t=11$ & The end card gives only the month-level date ``PART 1: NOVEMBER 2010,'' while external retrieval provides exact dates such as birth dates. A \texttt{temporal\_mismatch} conflict enters $C_t$. & 1 & No; an unresolved conflict remains.\\
$t=11$ (same step) & The critic rewrites two need targets: check whether the end card contains a full calendar date, and resolve the conflicting subtitle statements ``famously played by Kenneth Branagh'' and ``played by David Tennant.'' & 1 & No.\\
$t=12$--14 & Reinspection confirms that the end card is precise only to the month, which does not conflict with an exact external date, and the conflict is resolved. The critic repeatedly marks the actor attribution as CONFLICTING; after two targeted checks, it accepts Kenneth Branagh, supported by external birth-date records. & 0 & Yes.\\
After $t=14$ & Submit ``29 years, 4 months, and 5 days,'' which matches the reference answer. & 0 & Submit.\\
\bottomrule
\end{tabular}
\end{table}

\section{Implementation Details}
\label{sec:appendix-implementation}

\subsection{Tool configuration}

Table~\ref{tab:appendix-tools} lists the tools available to the planner, the function of each, and the asset types it applies to.

\begin{table}[htbp]
\caption{Tool inventory and asset availability.}
\label{tab:appendix-tools}
\centering\small
\begin{tabular}{@{}p{0.34\linewidth}p{0.42\linewidth}p{0.15\linewidth}@{}}
\toprule Tool & Function & Assets\\ \midrule
\texttt{subtitle\_grounding\_tool} & Ground question-relevant content with subtitle or transcript spans. & Video, audio\\
\texttt{audio\_scout\_tool} & Retrieve speech and non-speech audio events. & Video, audio\\
\texttt{clip\_grounding\_tool} & Retrieve the most relevant candidate clip intervals from a caption library. & Video\\
\texttt{frame\_confirm\_tool} & Sample candidate intervals for confirmation by the visual backend; supports multiple intervals in one call and an image-inspection mode. & Video, image\\
\texttt{web\_search\_tool} & Retrieve external public facts that complete media evidence. & Global\\
\texttt{code\_executor\_tool} & Run short deterministic Python computations for arithmetic or dates. & Global\\
\texttt{finish} & Submit the final answer. & Global\\
\bottomrule
\end{tabular}
\end{table}

\subsection{Inference-loop configuration}

Section~\ref{sec:experiments-setup} specifies the model backends and evaluation protocol; the exact versions are \texttt{gpt-5.2-2025-12-11} for GPT-5.2 and \texttt{gemini-3.1-pro-preview} for Gemini-3.1-Pro. This section records the remaining runtime constants and differences between experimental arms. All runs use at most 15 steps. Each planner call emits at most one tool call or one finalization attempt. The runtime first normalizes each tool response and matches it to an evidence need. When evidence-level validation is required, the reducer records the response together with the critic verdict. Every \texttt{finish} attempt undergoes an answer-level check. If the check fails, the reducer writes the diagnosis of missing evidence or conflict back to the ledger and the loop continues.

Each backend replacement arm replaces only one backend named in Table~\ref{tab:backend-sensitivity}; the arm that replaces both backends replaces two. All other tools, prompts, budgets, and evaluation settings remain the same as in the default configuration.

The state-ablation arms share the default backends, tool interface, step limit, and budget (Table~\ref{tab:state-ablation}); they differ only in the form in which prior observations reach the planner. No-ledger ReAct removes the ledger machinery entirely: each raw tool observation is appended chronologically to the dialogue history, and the planner reads this history to select the next action. With no structured state to check against, this arm has no readiness check or answer verification, and the planner decides on its own when to answer. Memory maintains context in the style of a memory plugin: after each tool return, an additional LLM call merges the new observation into a rolling summary that records confirmed facts, unfinished goals, and whether an answer can be attempted, and the planner reads only this summary at each step. The summary length is capped by the rendered ledger text of the full method at the same step, so the two arms read comparable amounts of information per step. All other components of the Memory arm, including critic validation, event commits, and stopping, match the full method. WorldSense runs disable web search and external fact completion, and accuracy is computed by matching the selected answer option.

\subsection{Web-search domain blocking and same-window comparison}
\label{sec:appendix-web-comparison}

The Omni-Decision web-search tool blocks Hugging Face domains to prevent retrieval of the benchmark data itself. In the comparison within the same window, Gemini-3.1-Pro answers questions directly with its built-in retrieval capability, to which we cannot apply the same block. Spot checks of its answer traces show clear signs of retrieval from Hugging Face. We report its measured score without adjustment.

\section{Failure Analysis}
\label{sec:appendix-failure}

\subsection{Evidence-progress protocol}

The ledger declares evidence needs for each task and records their closure. Figure~\ref{fig:progress-audit}(a) uses these records to measure closure of declared needs across all 360 questions. Correct cases close 66.2\% of their needs on average, compared with 32.8\% for incorrect cases. Among the incorrect cases, 92.5\% close at least one need.

\subsection{Failure localization}

For each of the 67 incorrect cases, we manually identify the earliest decisive failure in the evidence chain. Figure~\ref{fig:progress-audit}(b) shows the category distribution, and Table~\ref{tab:failure-localization} gives the counts.

\begin{table}[htbp]
\caption{Earliest decisive failure among 67 incorrect cases.}
\label{tab:failure-localization}
\centering\small
\begin{tabular}{lrr}
\toprule Stage & Cases & Share\\ \midrule
Media perception & 27 & 40.3\%\\
External-fact retrieval & 22 & 32.8\%\\
Verification / conflict resolution & 9 & 13.4\%\\
Computation & 6 & 9.0\%\\
Premature stopping / readiness & 3 & 4.5\%\\
\bottomrule
\end{tabular}
\end{table}

We assign a mutually exclusive primary cause to the earliest deviation that is not repaired later and propagates to the final answer. A media perception failure occurs when low-level reading misses required media evidence, such as small text, a fine-grained attribute, or a dense action, so the evidence never enters $E_t$. An external retrieval failure either misses the required fact or retrieves a fact for the wrong entity. A verification or conflict resolution failure omits a conflict that should enter $C_t$ or treats insufficient evidence as closing a need. A computation failure breaks the numerical or logical derivation. Premature stopping occurs when the system finalizes an answer or abstains before the ledger is ready.

\subsection{Two failure cases}

\paragraph{OmniGAIA \#68 (Geography \& Travel).}
The task requires reading a station name from an image, combining it with the departure port mentioned in the audio, and computing the distance between the two locations. The station name is the entry slot for the entire evidence chain. The run follows \texttt{frame\_confirm $\to$ subtitle\_grounding $\to$ frame\_confirm $\to$ audio\_scout $\to$ frame\_confirm $\to$ web\_search$\times2$ $\to$ finish (blocked) $\to$ web\_search $\to$ code\_executor$\times2$ $\to$ finish (forced)}. At the first \texttt{finish} attempt, $E_t$ still lacks a reliable station name, so the critic blocks finalization. Here, a forced answer is an undesirable exit after repeated blocks exhaust the revision budget, distinct from a normal answer that passes the check. Subsequent web searches and computations cannot replace the missing visual reading because, without the entry entity, they can operate only on an incorrect or empty object. The final output states that the station name is unreadable and gives an estimate of approximately 4.8 km, which is judged incorrect. The ledger localizes the failure to the unclosed station-name entry slot.

\paragraph{WorldSense \#1097 (Performance).}
The question asks what a man in white did to a machine gun. The missing evidence is a fine-grained action between the person and object. The run follows \texttt{clip\_grounding $\to$ frame\_confirm (multiple calls) $\to$ subtitle\_grounding $\to$ audio\_scout $\to$ clip\_grounding $\to$ frame\_confirm $\to$ finish (blocked) $\to$ finish (forced)}. The system locates the relevant person and object, but the action evidence remains unstable, and the critic blocks finalization twice. Further clips and frames still do not provide enough action evidence to distinguish the options. The final output selects C, while the reference answer is D.

In both cases, the ledger identifies a specific open slot at the time of failure, namely the entry evidence for the station name in the first case and the action evidence in the second, but subsequent tools do not fill it.

\section{Native and Toolized Omni-modal Backends}
\label{sec:appendix-backends}

Omni-Decision supports two interfaces to the perception layer: a native omni-modal backend that consumes the complete multimodal input directly, and an omni-modal backend invoked through perception tools according to ledger needs. The comparison in the original OmniGAIA paper shows a limited gap between these interfaces (Table~\ref{tab:native-toolized}). Gemini-3-Flash scores 51.7 with native input, 50.0 when only audio is toolized, a decrease of 1.7 points, and 46.4 when both audio and visual perception are toolized, a decrease of 5.3 points. The weaker Qwen3-Omni scores 13.3 with native input, while the toolized settings reach 15.8 to 18.1. Toolization can therefore offload some low-level reading and on-demand retrieval from a weaker model.

\begin{table}[htbp]
\caption{Native and toolized perception on OmniGAIA, excerpted from the benchmark paper \citep{liOmniGAIANativeOmniModal2026}.}
\label{tab:native-toolized}
\centering\scriptsize
\resizebox{\linewidth}{!}{%
\begin{tabular}{l@{\hspace{14pt}}rrr@{\hspace{14pt}}rr}
\toprule
\multirow{2}{*}{Model / setting} & \multicolumn{3}{@{}c@{\hspace{14pt}}}{Difficulty} & \multirow{2}{*}{Avg.} & \multirow{2}{*}{Avg. tool calls}\\
\cmidrule(l{0.2em}r{1.45em}){2-4}
 & Easy & Medium & Hard & & \\
\midrule
Gemini-3-Flash, native omni-modal &67.2&46.9&37.2&51.7&4.4\\
Gemini-3-Flash, visual native; audio toolized &60.7&48.8&35.9&50.0&7.6\\
Gemini-3-Flash, audio and visual toolized &52.5&46.9&35.9&46.4&9.4\\
Qwen3-Omni, native omni-modal &19.7&10.6&9.0&13.3&0.2\\
Qwen3-Omni, visual native; audio toolized &24.6&15.0&3.9&15.8&0.8\\
Qwen3-VL + Qwen3-Omni, visual toolized &24.6&18.1&7.7&18.1&2.8\\
Qwen3 + Qwen3-Omni, audio and visual toolized &32.8&10.6&6.4&17.2&2.3\\
\bottomrule
\end{tabular}}
\end{table}

Whether an observation comes from native input or a tool call, planning reads the same control projection of the ledger. Toolization changes how observations enter the system without changing the ledger update or decision procedure.

\section{Cost and Token Statistics}
\label{sec:appendix-cost}

\begin{table}[htbp]
\caption{Per-question cost from runtime billing records.}
\label{tab:appendix-cost}
\centering
\small
\begin{tabular}{llrrr}
\toprule Benchmark & System & Accuracy & Range & Mean\\ \midrule
OmniGAIA & \textbf{Omni-Decision} & \textbf{81.39} & \$0.46--\$2.77 & $\approx$\$1.2\\
OmniGAIA & Gemini-3.1-Pro & 79.44 & \$0.93--\$5.83 & $\approx$\$2.8\\
WorldSense & \textbf{Omni-Decision} & \textbf{65.00} & \$0.08--\$0.40 & $\approx$\$0.3\\
WorldSense & Gemini-3.1-Pro\textsuperscript{\emph{a}} & 65.50 & \$0.30--\$1.07 & $\approx$\$0.8\\
\bottomrule
\end{tabular}
\vspace{2pt}

\parbox{\linewidth}{\scriptsize \textsuperscript{\emph{a}}The WorldSense accuracy is the public leaderboard result; the cost is our runtime billing measurement for direct answers to the same questions. Costs in all rows are billing records and are not inferred from token counts.}
\end{table}

\begin{figure}[htbp]
  \centering
  \includegraphics[width=0.6\linewidth]{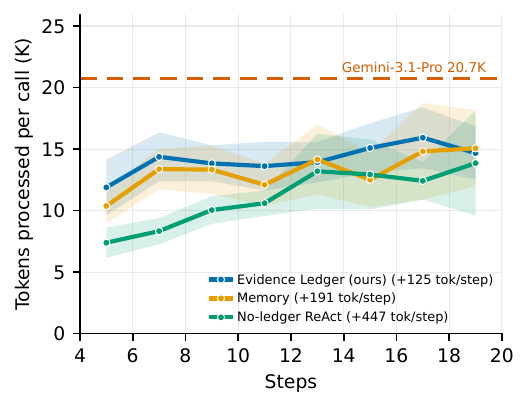}
  \caption{Tokens processed per planner call over execution steps for the three state-ablation arms, restricted to the step range shared by all arms. The dashed line is the average token count for a single direct Gemini-3.1-Pro call on the same 262-question pool.}
  \label{fig:backend-tokens}
\end{figure}

\paragraph{Token usage.}
The ledger's persistent context contains only the control projection, so prompt tokens per step remain approximately flat. Figure~\ref{fig:backend-tokens} uses the 262-question multi-asset pool shared by the three state-ablation arms to plot tokens processed per planner call, together with the 20.7K-token reference for one direct Gemini-3.1-Pro call on the same pool. All three arms stay below this reference because each per-step call omits the full media context. Over the full pool, direct Gemini-3.1-Pro answering uses one call and approximately 26.3K prompt tokens per example, while Omni-Decision uses approximately 29--31 calls and 204K prompt tokens. The system therefore processes more tokens in total, yet incurs the lower billed cost in Table~\ref{tab:appendix-cost}, because billed cost depends on the unit price of each token rather than on the raw count. Three factors separate the two configurations. Most of the token volume passes through text-only planner, critic, and finalizer calls, which are priced well below multimodal calls that ingest media. The perception backend reads only the clips and frames that open needs request, whereas a direct answer must ingest the complete media input. Finally, the persistent context is a short, stable control projection, so consecutive calls share their prompt prefix and hit the provider cache, whose tokens are billed at a discounted rate.

\section{Training Details}
\label{sec:appendix-training}

This appendix describes the data construction, training configuration, and reward used to post-train two planner backbones, Qwen3.5-27B and Qwen3-Omni-30B, on Omni-Decision trajectories.

\subsection{Training-data construction}
\label{sec:appendix-training-data}

\paragraph{Source.}
The default system configuration, with a GPT-5.2 planner and Gemini-3.1-Pro perception, generates the training trajectories by running the complete system. We retain only trajectories that the judge marks as correct. The runtime finish gate ensures that these trajectories terminate after evidence closure, and tool calls use the native function calling interface with runtime format validation. No additional trajectory cleaning is required.

\paragraph{Expansion into supervised pairs.}
We retain only the planner's decision steps from each trajectory. The input of a supervised pair is the complete conversation that the planner actually reads at that step, including the ledger text rendered into the observation. The target is the action produced at that step, including the tool choice, arguments, and termination decision. The training set contains 640 trajectories from two sources: 100 OmniGAIA Easy trajectories and 540 WorldSense trajectories. The latter are the subsets of two teacher runs that the judge marks correct, containing 288 and 252 trajectories. WorldSense trajectories use a tool set without web access, whereas OmniGAIA trajectories use the complete tool set with web access. Supervised pairs expanded from these 640 trajectories train both the Qwen3.5-27B and Qwen3-Omni-30B planners, and the states at their decision steps also form the reinforcement-learning state pool in Section~\ref{sec:appendix-training-rl}.

\paragraph{Relation between training and evaluation.}
The Medium and Hard evaluation sets, containing 160 and 78 questions, are disjoint from the training data by task ID. The Easy evaluation set and training trajectories draw from the same task source, but supervision consists of the teacher's action at each step and contains no answer labels. On Easy, state-SFT answers four additional questions correctly, increasing accuracy from 62.30\% to 65.57\%, a gain on the same scale as that on the fully held-out Medium set. Evaluation otherwise matches the corresponding rows of the untrained planners in Table~\ref{tab:backend-sensitivity}: it uses the trained planner with Gemini-3.1-Pro perception and the same tools and budgets, allowing direct row-wise comparison.

\subsection{Supervised fine-tuning (state-SFT)}
\label{sec:appendix-training-sft}

Both backbones use the same configuration: attention-only LoRA targeting \texttt{q/k/v/o\_proj}, bf16 precision, and sequence-parallel training for long trajectories. The visual encoder is frozen for the multimodal backbone. After training, the LoRA weights are merged into each backbone for evaluation. The merged Qwen3.5-27B model is also the initial policy and KL reference policy for reinforcement learning.

\subsection{Closure-aligned reinforcement learning}
\label{sec:appendix-training-rl}

\paragraph{Design basis.}
The framework's failure diagnosis and state structure determine the RL target and reward. First, Section~\ref{sec:experiments-backend} shows that failures of weak planners are dominated by decision-level behavior: the planner abstains or converges prematurely while evidence remains open, even though it receives ledger states and perceptual observations from the same sources as the default configuration. Optimizing individual decisions therefore targets the observed point of failure. Second, answer-level rewards require a complete system execution for every training sample, including multiple planner calls and perception and retrieval APIs. The signal is sparse, and cost grows with trajectory length. Because the ledger maintains an explicit list of open needs, it can evaluate a single decision without executing that action (Section~\ref{sec:method-training}). This approach and DA-GRPO in Orchestra-o1 both perform offline decision-level RL. DA-GRPO asks a judge to evaluate a rubric over the complete unstructured subtask history. In our four-dimensional reward, ledger rules directly determine two dimensions, and the ledger's list of open needs grounds the other two, reducing the reward judge's discretion.

\paragraph{State pool and reference decisions.}
For every decision step, we reconstruct the input state from the teacher trajectories that the judge marks correct (Section~\ref{sec:appendix-training-data}). Each state contains the same conversation and ledger rendering used for SFT, and the teacher's actual action at that step is attached as the reference decision. The state pool reuses the states at decision steps expanded from the same 640 trajectories as SFT, with the same relationship to evaluation described above.

\paragraph{Rubric reward.}
For each state $s_i$, the current policy samples $G$ candidate decisions $\{y_{ij}\}_{j=1}^{G}$. Each candidate receives the weighted four-dimensional score
\begin{equation}
\label{eq:closure-aligned-reward}
r_{ij} = \alpha_1 r^{\mathrm{valid}}_{ij}
       + \alpha_2 r^{\mathrm{gate}}_{ij}
       + \alpha_3 r^{\mathrm{target}}_{ij}
       + \alpha_4 r^{\mathrm{align}}_{ij}.
\end{equation}
The first two dimensions are binary rule-based scores computed directly from the tool schema and ledger, without a model. The action validity score $r^{\mathrm{valid}}$ checks that the action is parseable, the selected tool exists, and its arguments satisfy the schema. The termination consistency score $r^{\mathrm{gate}}$ is zero when the candidate selects \texttt{finish} while the current ledger still contains an open blocking need or unresolved conflict, and one otherwise. It uses the same readiness predicate as the runtime finish gate.

The remaining two dimensions are graded scores in $[0,1]$, produced together by a lightweight LLM reward judge in one call. The action targeting score $r^{\mathrm{target}}$ measures which open need the selected tool and arguments address and whether the resulting observation could advance that need toward closure. The judge bases this score on the ledger's list of open needs. The policy alignment score $r^{\mathrm{align}}$ uses the reference decision as an anchor and assigns a high score to the same action or a reasonable alternative path. A \texttt{finish} action receives full credit when the evidence is closed and the draft answer matches the ground truth. The judge receives the question, ground truth, current ledger state, reference decision, and candidate decision. This reward judge is independent of the official OmniGAIA evaluation judge, and the evaluation protocol does not enter the training signal.

The four dimensions cover the decision-level failure surface. The gate score targets the dominant behavior observed in Section~\ref{sec:experiments-backend}, namely abstention or premature convergence before evidence closure. The targeting and alignment scores cover mismatches between actions and needs and broader policy degradation, while the validity score prevents RL updates from damaging the action format.

\paragraph{Optimization objective.}
We normalize the $G$ candidates within each state to obtain the relative advantage
\begin{equation}
\label{eq:closure-aligned-advantage}
\widehat{A}_{ij}
= \frac{r_{ij} - \operatorname{mean}_{j'=1}^{G}(r_{ij'})}
       {\operatorname{std}_{j'=1}^{G}(r_{ij'}) + \epsilon}.
\end{equation}
The other candidates in the group provide the baseline, removing difficulty differences across states and retaining only relative differences among candidates, following GRPO and the decision-level formulation in Orchestra-o1 \citep{shaoDeepSeekMathPushingLimits2024,zhangOrchestraO1Omnimodal2026}. We update the policy with a clipped importance ratio policy gradient objective and apply KL regularization against the post-SFT initial policy, preventing RL from disrupting the action format and decision style learned during SFT. This procedure gives dense feedback to the planner's three central decisions: tool selection, argument generation, and termination. The reward is based on the state at a single step instead of the final answer; Section~\ref{sec:experiments-training} tests whether this proxy produces task-level gains.

\paragraph{Training configuration.}
We use a rollout group size of $G=8$, a learning rate of $5\times10^{-6}$ with cosine decay, a KL coefficient of $0.01$, and a clipping parameter of $\epsilon_{\mathrm{clip}}=0.2$, training for 5 epochs. The reward weights $(\alpha_1,\alpha_2,\alpha_3,\alpha_4)$ are $(0.1,0.1,0.2,0.6)$.

\section{Prompt Templates}
\label{sec:appendix-prompts}

The following five templates are reproduced from measured runs, with the runtime's internal system name replaced by Omni-Decision: Planner System, Planner User, Evidence Critic, Answer Critic, and Answer/Finalizer. Runtime fields such as \texttt{\{asset\_manifest\}} and \texttt{\{question\}} are filled for each example. The template field \texttt{evidence\_state\_digest} is the runtime text rendering of the ledger described in Section~\ref{sec:method-control}. Tool schemas are passed through an OpenAI-compatible function-calling interface.
\begin{prompttemplatebox}{Planner System}
\begin{PromptVerbatim}
You are Omni-Decision, a long-video multimodal QA runtime built around an explicit
Evidence State.

Your job is to run one state-driven loop:
evidence state digest -> choose one tool -> observe -> read critic checkpoint
-> continue or answer

Available tools:
- subtitle_grounding_tool -- Ground the question using subtitle or transcript
  segments.
- audio_scout_tool -- Ground audio-related questions via speech retrieval or
  non-speech audio retrieval.
- clip_grounding_tool -- Retrieve relevant clip-level candidate ranges.
- frame_confirm_tool -- Visually confirm candidate ranges by sampling frames and
  asking the VLM. It supports merge_ranges=true for multiple labeled ranges.
- web_search_tool -- Search the public web for external facts that supplement
  media evidence.
- code_executor_tool -- Execute deterministic Python for arithmetic, date math,
  and similar calculations.
- finish -- Submit the final answer and end the conversation.

For multiple assets, the user message provides an asset manifest. You MUST
specify asset_id for asset-specific tools. For images, frame_confirm_tool uses
image-inspection mode and does not accept time ranges.

Tool behavior suggestions:
- Prefer video-grounding tools before external search.
- Use web_search_tool after media evidence identifies the entity and the
  remaining gap is an external public fact.
- Use frame_confirm_tool for visual verification; use merge_ranges=true for
  comparisons across ranges or holistic scene understanding.
- Use code_executor_tool whenever deterministic computation is required.
- Call only one tool at a time.

When choosing the next tool:
1. Check evidence_needs and actionable_need_indices.
2. Prefer actionable needs over blocked_need_indices.
3. Match preferred_source to the relevant tool.
4. Check tool_call_counts and explore an untried required modality.
5. Use gap_diagnosis to identify missing or conflicting evidence.
6. For web search, combine query_focus with entities from filled anchor needs;
   do not search the entire question verbatim.

Evidence-gathering strategy:
- Use media grounding first to collect direct evidence.
- Guide anchor acquisition with query_focus.
- For external facts, combine query_focus with resolved anchor entities.
- Attempt compute needs only after their dependencies are filled.
- If the critic reports missing visual or audio evidence, explore media tools
  instead of repeating web search.
- If a path yields no evidence, try an unexplored modality or reformulate the
  search from a different angle.

Evidence State Digest:
- evidence_needs: requirements derived from the question, each with need,
  preferred_source, closure_level, status, filled_by, need_role, depends_on,
  and query_focus.
- actionable_need_indices: needs ready to pursue.
- blocked_need_indices: needs waiting on dependencies.
- tool_call_counts: tool-name to call-count mapping.
- gap_diagnosis: evidence present, missing, or conflicting.
- entity_cards: entities extracted from media evidence.
- fact_bridge_records: entity-specific external lookup results.
- finish_available: whether finish is currently allowed.

Precision policy:
- Do not infer exact dates, counts, numeric gaps, or specifications from vague
  evidence.
- Prefer the shortest faithful final answer.
- Treat a fact confirmed by multiple independent sources as established.

Before finish, verify that the evidence addresses the question, blocking needs
are closed unless finish_available is true, and the answer does not contradict
the ledger. If the answer checkpoint blocks, revise toward a specific supported
answer rather than a more hedged one.
\end{PromptVerbatim}
\end{prompttemplatebox}

\begin{prompttemplatebox}{Planner User}
\begin{PromptVerbatim}
Single-video template:
Answer this question about a video ({video_length} seconds long).

Question: {question}

Multi-asset template:
Answer this question about the following assets:

{asset_manifest}

Question: {question}
\end{PromptVerbatim}
\end{prompttemplatebox}

\begin{prompttemplatebox}{Evidence Critic}
\begin{PromptVerbatim}
You are the internal evidence critic for Omni-Decision.

You diagnose whether the evidence is sufficient. You do NOT route or suggest
tools. Output valid JSON only:
{
  "verdict": "SUFFICIENT or INSUFFICIENT or CONFLICTING",
  "gap_diagnosis": {
    "gap_type": "sufficient or insufficient or conflicting",
    "description": "one concise evidence diagnosis"
  },
  "updated_need_statuses": [
    {
      "need_index": 0,
      "status": "unfilled or partial or filled or unfillable",
      "filled_by": "tool_name or null",
      "revised_query_focus": "optional corrected search direction",
      "revised_need_role": "optional corrected role",
      "add_depends_on": "optional dependency indices"
    }
  ]
}

Use the stable zero-based array position as need_index. Do not rewrite the need.

Verdicts:
- SUFFICIENT: accumulated evidence supports a faithful answer. Indirect sources
  and multiple agreeing web sources can be sufficient.
- INSUFFICIENT: a critical answer fact is absent, not merely absent from an
  official source.
- CONFLICTING: evidence atoms contradict one another on an answer-critical fact.

For every evidence need, compare evidence_state_digest,
accumulated_observations, and latest_observation. Set its status to unfilled,
partial, filled, or unfillable; record filled_by; correct query_focus or role
when needed; add newly discovered dependencies; and return the full status list.
Mark repeated unsuccessful needs unfillable. A filled need must lock the exact
entity or value at sufficient precision; partial means relevant but imprecise.

If any critical need or an upstream dependency of a computation remains open,
the verdict is INSUFFICIENT. CONFLICTING overrides INSUFFICIENT for incompatible
concrete values. Formatting differences and values for different needs are not
conflicts. Treat evidence_state_digest as the primary authority. The diagnosis
must state exactly what is present, missing, or conflicting without naming a
tool, and should use tool_call_counts to assess explored modalities.
\end{PromptVerbatim}
\end{prompttemplatebox}

\begin{prompttemplatebox}{Answer Critic}
\begin{PromptVerbatim}
You are a final-answer quality checker for a video QA agent.

Output JSON only:
{"verdict": "PASS or BLOCK", "revision_hint": "..."}

PASS when proposed_answer is a concrete response that directly answers the
question and is supported by or reasonably derived from accumulated evidence.
An approximate best-effort answer may pass if it commits to a value and does not
contradict evidence.

BLOCK when the answer is empty, refuses or abstains, never commits to a value,
hedges without committing, or directly contradicts an evidence atom.

Do not block minor wording issues or evidence merely because it came from web
search. If a specific value is supported by an evidence atom, pass even when
system metadata says evidence is insufficient. Approximate computations may pass
when no contradicting evidence exists.

revision_hint when BLOCK: "Commit now. State only your best value from the evidence."
revision_hint when PASS: ""
\end{PromptVerbatim}
\end{prompttemplatebox}

\begin{prompttemplatebox}{Answer / Finalizer}
\begin{PromptVerbatim}
You are a concise QA assistant. Based on the available evidence, give the best
direct answer to the question. If a computation result exists in the evidence,
use it directly. Respond with the answer only.

Input:
- Question: {question}
- Evidence summary: {answer_phase_evidence_summary}

Output:
- the shortest answer string compatible with the benchmark answer format.
\end{PromptVerbatim}
\end{prompttemplatebox}
\end{document}